\documentclass[11pt,a4paper]{article}
\usepackage{authblk}
\usepackage[hyperref]{acl2018}
\usepackage{times}
\usepackage{latexsym}

\usepackage{url}

\usepackage{booktabs} 
\usepackage{subcaption} 
\usepackage{verbatim} 
\usepackage{enumitem} 

\usepackage{todonotes}

\newcommand{\0}{\hspace*{0.5em}}

\aclfinalcopy 

\title{ScoutBot: A Dialogue System for Collaborative Navigation}

\author{Stephanie M. Lukin$^1$, Felix Gervits$^2$\thanks{\0Contributions were primarily performed during an internship at the Institute for Creative Technologies.}\0, Cory J. Hayes$^1$, Anton Leuski$^3$, Pooja Moolchandani$^1$, \authorcr John G. Rogers III$^1$, Carlos Sanchez Amaro$^1$, Matthew Marge$^1$, Clare R. Voss$^1$, David Traum$^3$ \\
$^1$ U.S. Army Research Laboratory, Adelphi, MD 20783\\
$^2$ Tufts University, Medford MA 02155\\
$^3$ USC Institute for Creative Technologies, Playa Vista, CA 90094\\
\textit {stephanie.m.lukin.civ@mail.mil, felix.gervits@tufts.edu, traum@ict.usc.edu}}

\date{}

\begin{document}
\maketitle
\begin{abstract}
ScoutBot is a dialogue interface to physical and simulated robots
that supports collaborative exploration of environments.
The demonstration will allow users to issue unconstrained spoken
language commands to ScoutBot. ScoutBot will prompt for clarification
if the user's instruction needs additional input. 
It is trained on human-robot dialogue collected from
Wizard-of-Oz experiments, where robot responses were initiated
by a human wizard in previous interactions. The demonstration
will show a simulated ground robot (Clearpath Jackal)
in a simulated environment
supported by ROS (Robot Operating System).
\end{abstract}

\section{Introduction}

\begin{figure}[t!]
    \centering
    \begin{subfigure}[t]{0.2\textwidth}
        \centering
        \includegraphics[width=0.8in]{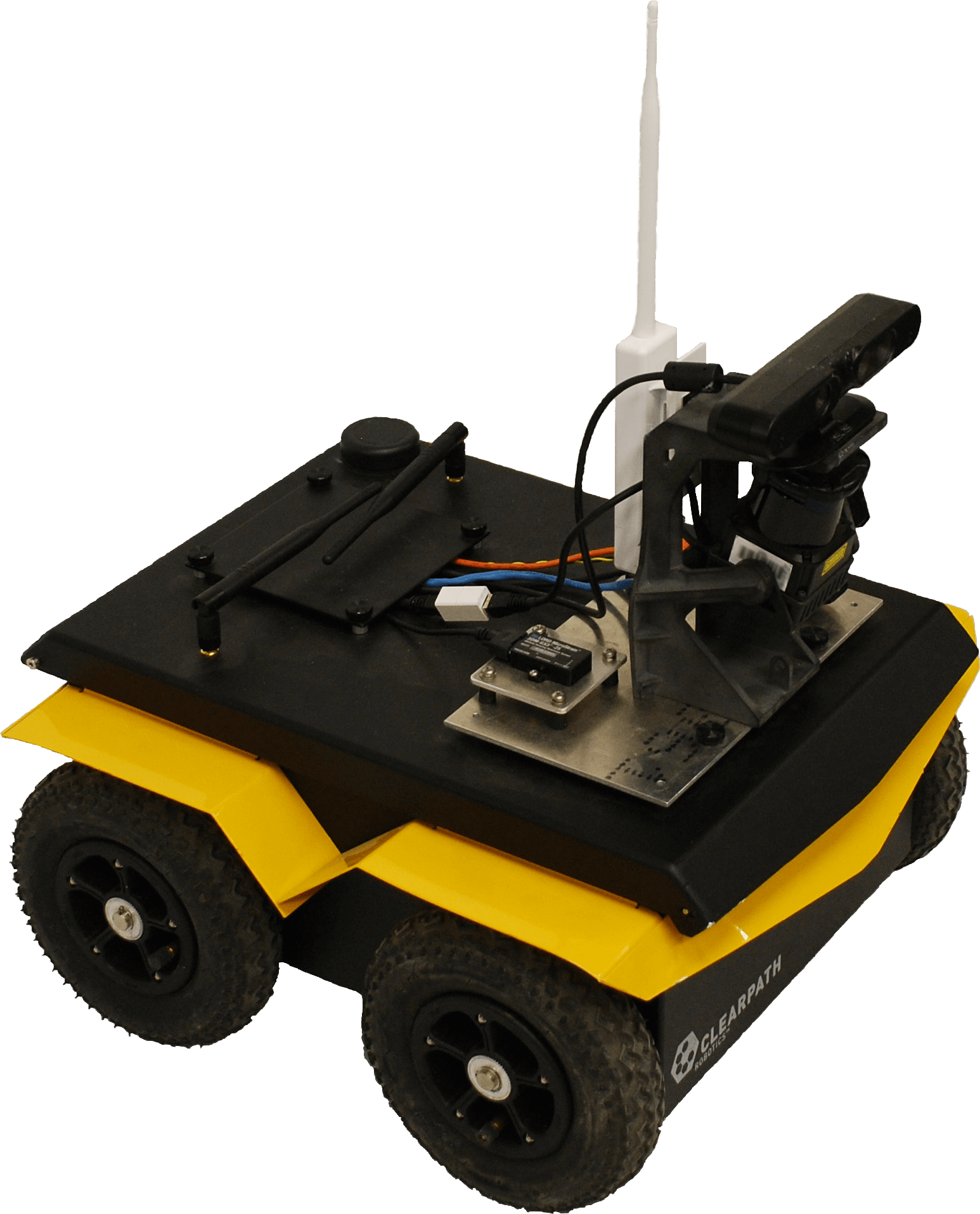}
        \caption{\label{jackal-real}Real-World Jackal}
    \end{subfigure}%
    ~ 
    \begin{subfigure}[t]{0.2\textwidth}
        \centering
        \includegraphics[width=0.8in]{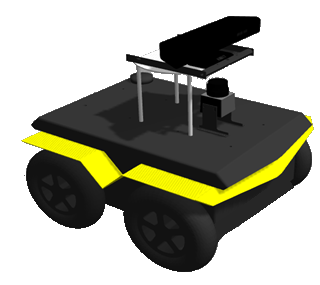}
        \caption{\label{jackal-virtual}Simulated Jackal}
    \end{subfigure}
    \caption{Jackal Robot}
\vspace{-0.2in}
\end{figure}

We are engaged in a long-term project to create an intelligent,
autonomous robot that can collaborate with people in remote locations to explore the
robot's surroundings.  The robot's capabilities will enable it to
effectively use language and other modalities in a natural manner for dialogue with a human teammate.
This demo highlights the current stage
of the project: a data-driven, automated system, ScoutBot, that can control a
simulated robot with verbally issued, natural language instructions
within a simulated environment, and can communicate in a manner similar to
the interactions observed in prior Wizard-of-Oz experiments. We used a Clearpath Jackal robot (shown in Figure~\ref{jackal-real}), a small, four-wheeled unmanned ground vehicle with an onboard CPU and inertial measurement unit,
equipped with a camera and \underline{li}ght \underline{d}etection \underline{a}nd \underline{r}anging (LIDAR) mapping that readily allows for automation.
The robot's task is to navigate
through an urban environment (e.g., rooms in an apartment or an alleyway
between apartments), and communicate discoveries to a human
collaborator (the Commander). The Commander verbally provides
instructions and guidance for the robot to navigate the space.

The reasons for an intelligent robot collaborator, rather than one
teleoperated by a human, are twofold. First, the human resources
needed for completely controlling every aspect of robot motion
(including low-level path-planning and navigation) may not be
available.  Natural language allows for high-level tasking, specifying
a desired plan or end-point, such that the robot can figure out the
details of how to execute these natural language commands in the given context and report back to the
human as appropriate, requiring less time and cognitive load on
humans. Second, the interaction should be robust to low-bandwidth and
unreliable communication situations (common in disaster relief and
search-and-rescue scenarios), where it may be impossible or
impractical to exercise real-time control or see full video
streams. Natural language interaction coupled with low-bandwidth,
multimodal updates addresses both of these issues and provides less
need for high-bandwidth, reliable communication and full attention of
a human controller.

We have planned the research for developing ScoutBot as consisting of five conceptual phases,
each culminating in an experiment to validate the approach and collect
evaluation data to inform the development of the subsequent phase. These
phases are:

\begin{enumerate}
\small
\item{Typed wizard interaction in real-world environment} \vspace{-2ex}
\item{Structured wizard interaction in real environment}\vspace{-2ex}
\item{Structured wizard interaction in simulated environment}\vspace{-2ex}
\item{Automated interaction in simulated environment}\vspace{-2ex}
\item{Automated interaction in real environment}
\end{enumerate}

The first two phases are described in~\citet{marge2016assessing}
and~\citet{bonial2017laying_long}, respectively. Both consist of
Wizard-of-Oz settings in which participants believe that they are interacting with an autonomous robot, a common tool in Human-Robot Interaction for supporting not-yet fully realized algorithms \cite{riek2012wizard}. In our two-wizard design, one wizard (the Dialogue
Manager or DM-Wizard) handles the
communication aspects, while another (the Robot Navigator or RN-Wizard)
handles robot navigation.
The DM-Wizard acts as an interpreter between the Commander and robot,
passing simple and full instructions to the RN-Wizard for execution based on the Commander instruction
(e.g., the Commander instruction, {\it Now go and make a right um 45 degrees} is passed to the RN-Wizard for execution as, {\it Turn right 45 degrees}).
In turn, the DM-Wizard informs the Commander of instructions the RN-Wizard successfully executes or of problems that arise during execution.
Unclear instructions from the Commander are clarified through
dialogue with the DM-Wizard (e.g., {\it How far should I turn?}). Additional discussion between Commander and DM-Wizard is allowed at
any time. Note that because of the aforementioned bandwidth and
reliability issues, it is not feasible for the robot to start turning or moving and wait
for the Commander to tell it when to stop - this may cause the robot to
move too far, which could be dangerous in some circumstances and
confusing in others.
 In the first phase, the DM-Wizard uses
unconstrained texting to send messages to both the Commander and RN-Wizard. In the second phase, the DM-Wizard uses a click-button interface that facilitates
faster messaging. The set of DM-Wizard messages in this phase were
constrained based on the messages from the first phase. 

This demo introduces ScoutBot automation of the robot to be used in
the upcoming phases: a simulated robot and simulated environment to support the
third and fourth project phases, and initial automation of DM and RN roles, to be used
in the fourth and fifth phases. Simulation and automation were
based on analyses from data
collected in the first two phases.
Together, the simulated
environment and robot
allow us to test the automated system in a safe environment, where
people, the physical robot, and the real-world environment are not at risk due to
communication or navigation errors.

\section{System Capabilities}
ScoutBot engages in collaborative exploration dialogues with a
human Commander, and controls a robot to execute instructions and
navigate through and observe an environment. The real-world Jackal robot measures 20in x 17in x 10in, and weights about 37lbs (pictured in Figure~\ref{jackal-real}). Both it and its simulated counterpart (as seen in Figure~\ref{jackal-virtual}) can drive around the environment, but cannot manipulate objects or otherwise interact with the environment. While navigating, the robot uses LIDAR to
construct a map of its surroundings as well as to indicate its position in the
map. The Commander is shown this information, as well as
static photographs of the robot's frontal view in the environment (per
request) and dialogue responses in a text interface.

\begin{figure*}[t!]
 \centering
 \includegraphics[width=6in]{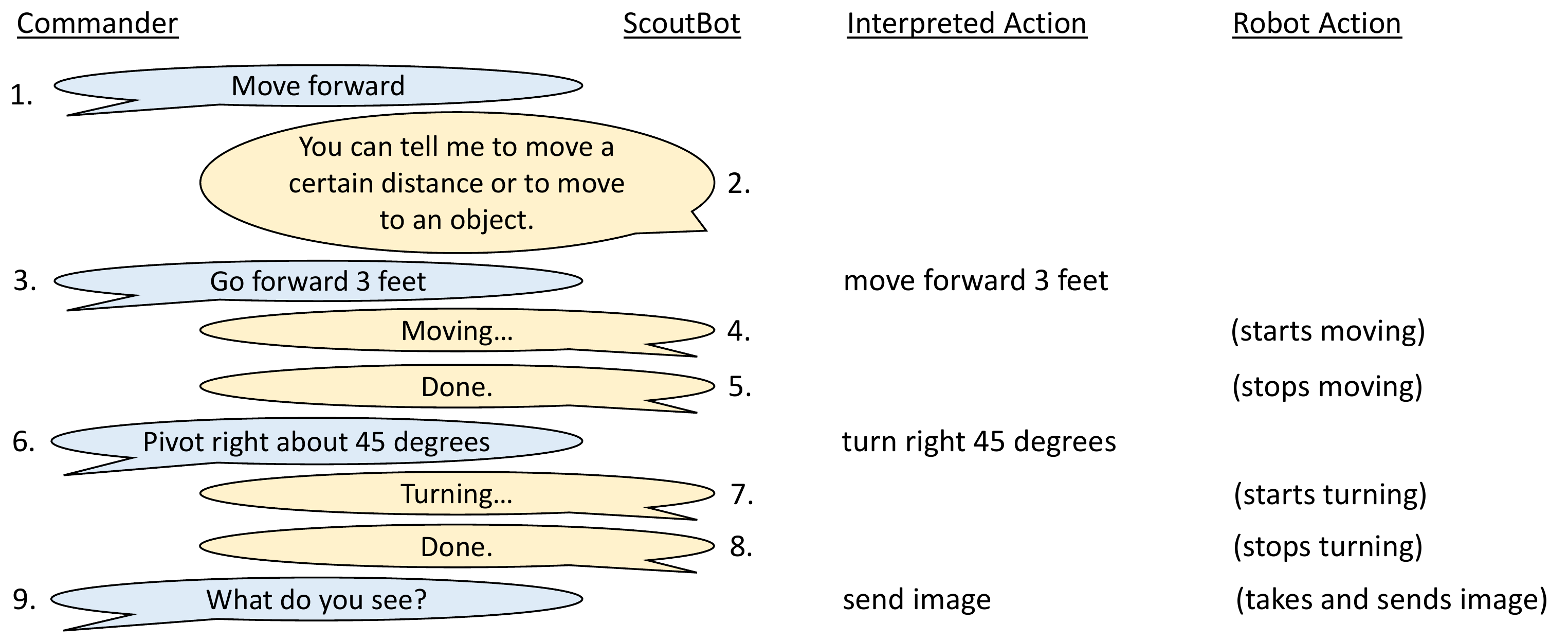}
\caption{Dialogue between a Commander and ScoutBot}
\vspace{-0.1in}
\label{fig:dialogue}
\end{figure*}

Interactions with ScoutBot are collaborative in that ScoutBot and the Commander exchange information through dialogue. Commanders issue spoken navigation commands to the robot, request photos, or ask questions
about its knowledge and perception. ScoutBot responds to indicate when commands are actionable, and gives status feedback on its
processing, action, or inability to act. When ScoutBot accepts a
command, it is carried out in the world. 

Figure~\ref{fig:dialogue} shows a dialogue between a Commander and ScoutBot. The right two columns
show how the Commander's language was interpreted, and the action the robot takes.
In this dialogue, the Commander begins by issuing a single
instruction, {\it Move forward} (utterance \#1), but the end-point is
underspecified, so ScoutBot responds by requesting additional
information (\#2). The Commander supplies the information in \#3, and
the utterance is understood as an instruction to move forward 3
feet. ScoutBot provides feedback to the Commander as the action is begins (\#4) and upon completion (\#5). Another successful action is executed in \#6-8. Finally, the Commander's request to know what the robot sees in \#9 is interpreted as a request for a picture from the robot's camera, which is taken and sent to the Commander.

ScoutBot accepts unconstrained spoken language, and uses a statistical
text classifier trained on annotated data from the first two phases of
the project for interpreting Commander instructions. Dialogue
management policies are used to generate messages to both the
Commander and interpreted actions for automated robot navigation. Our initial dialogue
management policies are fairly simple, yet are able to handle a wide
range of phenomena seen in our domain.

\section{System Overview}

\begin{figure}[t!]
 \centering
 \includegraphics[width=3in]{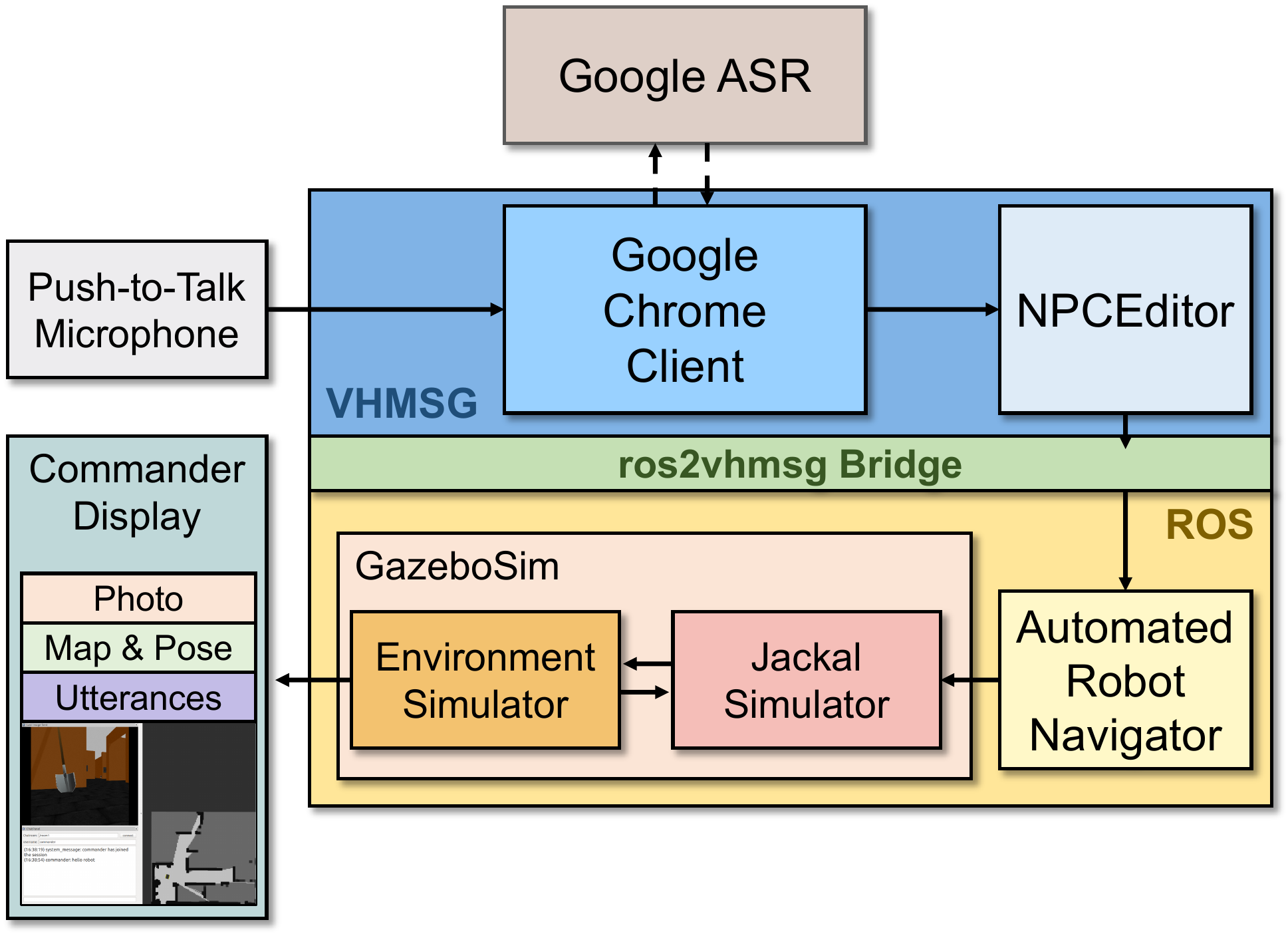}
 \vspace{-0.2in}
\caption{ScoutBot architecture interfacing with a simulated robot and environment:
Solid arrows represent
communications over a local network; dashed arrows
indicate connections with external resources. 
Messaging for the spoken language interface is handled via VHMSG, while robot messages are facilitated by ROS.
A messaging bridge, ros2vhmsg, connects the components.}
\vspace{-0.15in}
 \label{arch}
\end{figure}

ScoutBot consists of several software systems, running on multiple
machines and operating systems, using two distributed agent
platforms. The architecture utilized in simulation is shown in Figure~\ref{arch}. A parallel architecture exists for a real-world robot and environment. Components
primarily involving language
processing use parts of the Virtual Human (VH) architecture~\cite{hartholt2013},
and communicate using the Virtual Human Messaging (VHMSG), a thin layer
on top of the publicly available ActiveMQ message
protocol\footnote{\url{http://activemq.apache.org}}.  Components
primarily involving the real or simulated world, robot locomotion, and
sensing are embedded in ROS\footnote{\url{http://www.ros.org}}.
To allow the 
VHMSG
and ROS modules to interact with each other, we created ros2vhmsg,
software that bridges the two messaging architectures.
 The components are described in the remainder of this section.

The system includes several distinct workstation displays for
human participants. The Commander display is the view
of the collaborative partner for the robot (lower left corner of Figure~\ref{arch}). This display shows a map of the
robot's local surroundings, the most recent photo the robot
has sent, and a chat log showing text utterances from the robot. The map
is augmented as new areas are explored, and updated according
to the robot's
position and orientation.  There are also displays for experimenters to
monitor (and in the case of Wizards-of-Oz, engage in) the interaction. These displays show real-time
video updates of what the robot can see, as well as internal
communication and navigation aids.

\subsection{VHMSG Components}
VHMSG includes several protocols that implement parts of the Virtual
Human architecture. We use the protocols for speech recognition, as
well as component monitoring and logging. The NPCEditor and other
components that use these protocols are
available as part of the ICT Virtual Human Toolkit
\cite{hartholt2013}.
These protocols have also been used in systems, as reported by  \citet{hill-et-al-mre:2003} and \citet{Traum-et-al-twins2012,traum-EtAl:2015:Sigdial}. In particular, we used the 
adaptation of Google's Automated Speech Recognition (ASR) API
used in~\citet{traum-EtAl:2015:Sigdial}. The
NPCEditor~\cite{LeuskiTraum2011} was used for Natural Language Understanding (NLU) and dialogue
management. The new ros2vhmsg component for bridging the
messages was used to send instructions from the NPCEditor to the automated RN.

\subsubsection{NPCEditor}
We implemented NLU using the statistical text classifier
included in the NPCEditor. The classifier learns a mapping from inputs to
outputs from training data using cross-language retrieval models
\cite{LeuskiTraum2011}. The dialogues collected from our first two
experimental phases served as training data, and consisted of 1,500
pairs of Commander (user) utterances and the DM-Wizard's responses. While this approach limits responses to the set that were seen
in the training data, in practice it works well in our domain, 
achieving accuracy on a held out test set of 86\%.
The system is particularly effective at translating actionable commands to
the RN for execution (e.g., \textit{Take a picture}).
It is robust at handling
commands that include pauses, fillers, and other disfluent features
(e.g., \textit{Uh move um 10 feet}). It can also handle simple metric-based motion
commands (e.g., \textit{Turn right 45 degrees},
\textit{Move forward 10 feet})
as well as action sequences (e.g., \textit{Turn 180 degrees and take a
picture every 45}). 
The system can interpret some landmark-based instructions requiring knowledge
of the environment (e.g., \textit{Go to the orange cone}), 
but the automated RN component does not yet have the capability to generate the low-level, landmark-based instructions for the robot.

Although the NPCEditor supports simultaneous participation in multiple
conversations, extensions were needed for ScoutBot to support
multi-floor interaction~\cite{traum2018dialogue}, in which two
conversations are linked together. Rather than just responding to
input in a single conversation, the DM-Wizard in our first project phases
often translates input from one conversational floor to another (e.g., from the Commander to the RN-Wizard, or visa versa), or responds to input with
messages to both the
Commander and the RN. These responses need to be
consistent (e.g. translating a command to the RN should be accompanied
by positive feedback to the Commander, while a clarification to the
commander should not include an RN action command). Using the dialogue
relation annotations described in~\citet{traum2018dialogue}, we trained a
hybrid classifier, including translations to RN if they existed, and
negative feedback to the Commander where they did not. We also created
a new dialogue manager policy that would accompany RN commands with
appropriate positive feedback to the commander, e.g., response of
``Moving..'' vs  ``Turning...'' as seen in \#4 and \#7 in Figure~\ref{fig:dialogue}.

\subsubsection{ros2vhmsg}

ros2vhmsg is a macOS application to bridge the 
VHMSG 
and ROS components of ScoutBot. It can be run from a command line or
using a graphical user interface that simplifies the configuration of the application.  Both
VHMSG 
and ROS are publisher-subscriber architectures
with a central broker software. Each component connects to the
broker. They publish messages by delivering them to the broker and
receive messages by telling the broker which messages they want
to receive. When the broker receives a relevant message,
it is delivered to the subscribers.

ros2vhmsg registers itself as a client for both 
VHMSG 
and ROS brokers, translating and reissuing the messages. For example, when ros2vhmsg receives a message from the ROS broker, it converts the
message into a compatible 
VHMSG 
format and publishes it with the
VHMSG
broker. Within ros2vhmsg, the message names
to translate along with the corresponding ROS message types must be
specified.
Currently, the application
supports translation for messages carrying either text or robot
navigation commands. The application is flexible and easily extendable with additional message conversion rules.
ros2vhmsg annotates its messages with a special metadata symbol
and uses that information to avoid processing messages that it
already published.

\subsection{ROS Components}
\label{ROS}

ROS provides backend support for robots to operate in both real-world and simulated environments. Here, we describe our simulated testbed and automated navigation components. Developing automated components in simulation allows for a safe test space before software is transitioned to the real-world on a physical robot. 

\subsubsection{Environment Simulator}

\begin{figure}[t]
\centering
\includegraphics[width=1.49in]{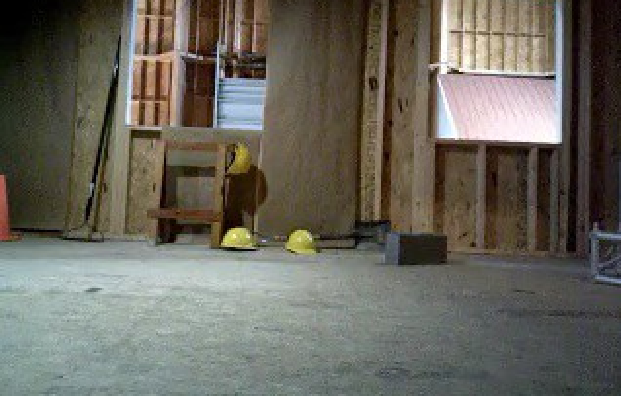}
\includegraphics[width=1.5in]{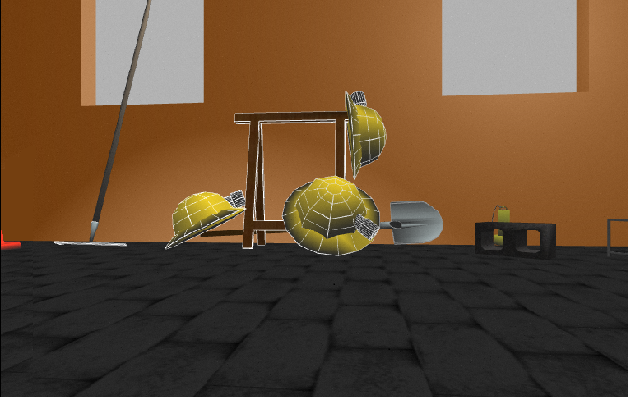}
\caption{Real-world and simulated instances of the same environment.}
\vspace{-0.1in}
\label{fig:env}
\end{figure}

Running operations under ROS makes the transition between
real-world and simulated testing straightforward. Gazebo\footnote{{\url{http://gazebosim.org/}}} is a software package
compatible with ROS for rendering high-fidelity
virtual simulations, and supports communications in the same manner
as one would in real-world environments.
Simulated environments were modeled in Gazebo after their real-world counterparts as shown in Figure~\ref{fig:env}. Replication
took into consideration the general dimensions of the physical space, and the location and size of objects that populate the space. Matching the realism and fidelity of the real-world environment in simulation comes with a rendering trade-off: objects requiring a higher polygon count due to their natural geometries results in slower rendering in Gazebo and a lag when the robot moves in the simulation. As a partial solution, objects were constructed starting from their basic geometric representations, which could be optimized accordingly, e.g., the illusion of depth was added with detailed textures or shading on flat-surfaced objects.
Environments rendered in Gazebo undergo a complex workflow
to support the aforementioned requirements. Objects and their vertices are placed in an environment, and properties about collisions and gravity are encoded in the simulated environment.

\subsubsection{Jackal Simulator}
Clearpath provides a ROS package with a simulated model of the Jackal robot (Figure~\ref{jackal-virtual}) and customizable features to create different simulated configurations. We configured our simulated Jackal to have the default inertial measurement unit, a generic camera, and a Hokuyo LIDAR laser scanner.

As the simulated Jackal navigates through the Gazebo environment, data
from the sensors is relayed to the workstation views through rviz\footnote{{\url{http://wiki.ros.org/rviz}}}, a visualization tool included with ROS that reads and displays sensor data in real-time.
Developers can select the data to display by adding a panel, selecting the data type, and then selecting the specific ROS data stream.
Both physical and simulated robot sensors are supported by the same rviz configuration, since rviz only processes the data sent from these sensors; this means that an rviz configuration created for data from a simulated robot will also work for a physical robot if the data stream types remain the same.

\subsubsection{Automated Robot Navigator}
Automated robot navigation is implemented with a python script and the {\sc rospy} package\footnote{{\url{http://wiki.ros.org/rospy}}}. The script runs on a ROS-enabled machine running the simulation. The script subscribes to messages from NPCEditor, which are routed through ros2vhmsg. These messages contain text output from the NLU classifier that issue instructions to the robot based on the user's unconstrained speech.

A mapping of pre-defined instructions was created, with keys matching the strings passed from NPCEditor via ros2vhmsg (e.g., {\it Turn left 90 degrees}). For movement, the map values are a ROS {\sc twist} message that specifies the linear motion and angular rotation payload for navigation. These {\sc twist} messages are the only way to manipulate the robot; measures in units of feet and degrees cannot be directly translated.
The mapping is straightforward to define for metric instructions. Included is basic coverage of moving forward or backward between 1 and 10 feet, and turning right and left 45, 90, 180, or 360 degrees. 
Pictures from the robot's simulated camera can be requested by sending a ROS {\sc image} message.
The current mapping does not yet support collision avoidance or landmark-based instructions that require knowledge of the surrounding environment (e.g., \textit{Go to the nearby chair}).

\section{Demo Summary}
In the demo, visitors will see the simulated robot dynamically move in the simulated environment, guided by natural language interaction. Visitors
will be able to speak instructions to the robot to move in the environment. Commands can be
given in metric terms (e.g., \#3 and \#6 in Figure~\ref{fig:dialogue}), and images requested (e.g., \#9).
Undecipherable or incomplete
instructions will result in clarification subdialogues rather than
robot motion (e.g., \#1). A variety of command formulations can be accommodated by
the NLU classifier based on the training data from our experiments.
Visualizations of different
components can be seen in:
{\url{http://people.ict.usc.edu/~traum/Movies/scoutbot-acl2018demo.wmv}}.

\section{Summary and Future Work}

ScoutBot serves as a research platform to support experimentation. 
ScoutBot components will be used in our upcoming third through fifth development phases. We are currently piloting phase 3 using ScoutBot's simulated environment, with human wizards. Meanwhile, we are extending the automated dialogue and navigation capabilities. 

This navigation task holds potential for collaboration policies to be studied, such as the amount and type of feedback given, how to negotiate to successful outcomes when an initial request was underspecified or impossible to carry out, and the impact of miscommunication. More sophisticated NLU methodologies can be tested, including those that recognize specific slots and values or more detailed semantics of spatial language descriptions. The use of context, particularly references to relative landmarks, can be tested by either using the assumed context as part of the input to the NLU classifier, transforming the input before classifying, or deferring the resolution and requiring
the action interpreter to handle situated context-dependent instructions \citep{kruijff2007incremental}.

Some components of the ScoutBot architecture may be substituted for different needs, such as different physical or simulated environments, robots, or tasks. Training new DM and RN components can make use of this general architecture, and the resulting components aligned back with ScoutBot.

\section*{Acknowledgments}
This work was supported by the U.S. Army.

\bibliography{acl2018demo}
\bibliographystyle{acl_natbib}


\end{document}